\documentclass{article}
\usepackage{soul}
\usepackage{array}
\usepackage{url}
\usepackage[utf8]{inputenc} \usepackage{hyperref}
\usepackage{CJKutf8}
\usepackage{graphicx} %fausto delete
\usepackage{amsmath} %fausto delete
\usepackage{latexsym}
\usepackage{booktabs}
\usepackage{balance}

\def\textasgraphics#1{\raisebox{-1.5mm}{\includegraphics[height=4.5mm]{#1}}}
\def\texthindi#1{\textasgraphics{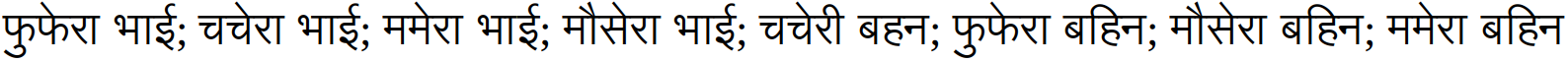}}
\def\textmalayalam#1{\raisebox{-1pt}{\includegraphics[height=3mm]{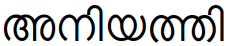}}}
\def\texttamil#1{\raisebox{-4pt}{\includegraphics[height=4mm]{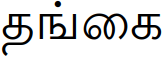}}}
\def\textchinese#1{\begin{CJK}{UTF8}{gbsn}#1\end{CJK}}
\def\para#1{\vspace{5pt}\noindent\textbf{#1} }

\begin{document}

\title{Representing Interlingual Meaning in Lexical Databases%\thanks{This work has been funded by the European Union project \texttt{InteropEHRate}, grant agreement number 826106.}
}

%\author*[1,2]{\fnm{Fausto} \sur{Giunchiglia}}\email{fausto.giunchiglia@unitn.it}
%\author*[1]{\fnm{Gábor} \sur{Bella}}\email{gabor.bella@unitn.it}
%\author[1]{\fnm{Nandu} \sur{C.~Nair}}\email{nandu.chandrannair@unitn.it}
%\author[2]{\fnm{Yang} \sur{Chi}}\email{yangchi19@mails.jlu.edu.cn}
%\author[2]{\fnm{Hao} %\sur{Xu}}\email{xuhao@jlu.edu.cn}

\author{Fausto Giunchiglia,*$^\dagger$ Gábor Bella,* Nandu Chandran Nair,*\\Yang Chi,$^\dagger$ and Hao Xu$^\dagger$}
\date{*University of Trento, $^\dagger$Jilin University}

%\affil[1]{\orgdiv{Department of Information Engineering and Computer Science}, \orgname{University of Trento}, \orgaddress{\street{via Sommarive, 5}, \city{Trento}, \postcode{38123}, \country{Italy}}}
%\affil[2]{\orgdiv{College of Computer Science and Technology}, \orgname{Jilin University}, \orgaddress{\city{Changchun}, \country{China}}}

%\keywords{language diversity, lexical semantics, multilingual lexical database, untranslatability}

\maketitle

\begin{abstract}
In today's multilingual lexical databases, the majority of the world's languages are under-represented. Beyond a mere issue of resource incompleteness, we show that existing lexical databases have structural limitations that result in a reduced expressivity on culturally-specific words and in mapping them across languages. In particular, the lexical meaning space of dominant languages, such as English, is represented more accurately while linguistically or culturally diverse languages are mapped in an approximate manner. Our paper assesses state-of-the-art multilingual lexical databases and evaluates their strengths and limitations with respect to their expressivity on lexical phenomena of linguistic diversity.  %Introducing a new notion of \emph{structural bias}, we analyze in detail the expressive power of the cross-lingual mappings of the state-of-the-art multilingual lexical databases over a dataset of culturally diverse words. 
%We then describe the \emph{Universal Knowledge Core}, a diversity-aware large-scale lexical database with an internal structure that addresses the shortcomings of existing mapping models.
\end{abstract}

\section{Introduction}

According to the Ethnologue \cite{ethnologue}, there are around seven thousand languages actively spoken in the world today. Despite the immense value---cultural, communicational, economic, etc.---embedded in languages and dialects, whether living or ancient, most computational resources on language have so far focused on a small subset of them, namely those spoken in the richest parts of the world \cite{Henrich2015}. Suggestive studies from \cite{kornai2013digital} and \cite{divide} articulate very well how the digitally less favoured populations suffer from  what is called the \textit{Digital Language Divide}, in terms of linguistic and cultural impoverishment.
In particular, beyond single-language lexical resources, \emph{multilingual lexical databases} (MLDB) play a pivotal role in language technologies such as cross-lingual word sense disambiguation, machine translation, or multilingual language models. They are also crucial for endangered and minority languages: for putting them in relation with all the world's languages, as reference material for language learners, and as a kind of knowledge-driven technology that complements corpus-based approaches in the absence of large corpora. %It is not a coincidence that in 2022 the Association for Computational Linguistics (ACL) conference had a Theme Track on \textit{``Language Diversity: from Low-Resource to Endangered Languages.''}

The goal of this paper is to draw upon these needs and to assess the state of the art in the development of MLDBs. We consider this survey a first step to drive future efforts. The key issue on which we concentrate is that no two vocabularies represent the world in exactly the same way, due to the pervasiveness of diversity in language, culture, and in how reality is perceived differently around the world. MLDBs need to capture these differences in expressivity \cite{giunchiglia2017understanding,giunchiglia2018one} and deal with  \emph{untranslatability} and cross-lingual shifts of meaning \cite{catford1978linguistic}. A failure to represent the linguistically or culturally specific elements of the vocabulary of a language may lead to a \emph{loss of function} \cite{kornai2013digital} and to an imposed uniformization with the world's dominant languages \cite{bella2022linguistic}.
Our paper has two main contributions: 

\begin{itemize}
    \item  a qualitative analysis of state-of-the-art MLDBs, reviewed according to four criteria that together guarantee an unbiased and diversity-aware representation of interlingual meaning; and 
    \item a complementary, quantitative evaluation of interlingual representation ability of these MLDBs over a corpus of about two thousand gold-standard interlingual mappings from linguistically and culturally diverse lexical fields.
\end{itemize}
Our analysis makes evident the pervasiveness of lexical untranslatability---the impossibility to find suitable concise translations for a word in another language---and the lack of computational resources that provide such evidence. A second take-home message is that representing interlingual meaning is, before anything else, a problem of lexico-semantic knowledge \emph{structure}: the lexical model underlying a MLDB intrinsically constrains its capability to represent lexical diversity. To scale to all the world's languages, the model needs to be powerful enough to capture, at the very least, the interlingual correspondences used in traditional lexicography: \emph{equivalence} when words ``have the same meaning'' for practical purposes, \emph{broader--narrower} relationships, and in case of untranslatability, indicating the presence of a \emph{lexical gap} as well as suitable broader terms as alternatives. 

\newcolumntype{L}[1]{>{\raggedright\let\newline\\\arraybackslash\hspace{0pt}}m{#1}}
\begin{table}[t]
\footnotesize
\setlength{\tabcolsep}{3pt}
\begin{center}
\begin{tabular}{l|L{4.5cm}|L{4.4cm}}
\textbf{Type} & \textbf{Source meaning} & \textbf{Target meaning}\\\hline\hline
%\textbf{equivalence} & relative$_\text{ ENG}$ & person related by blood or marriage & 亲戚$_\text{CHI}$ & person related by blood or marriage \\\hline
\textbf{equivalence} & MALAYALAM~\textmalayalam{അനിയത്തി} younger~daughter~of~father's~sister & TAMIL~\texttamil{தங்கை} younger~daughter~of~father's~sister\\\hline
\textbf{untranslatability} & CHINESE~\textchinese{堂妹}\penalty-10000 younger female patrilineal cousin  & ENGLISH \emph{lexical gap} \\\hline
\textbf{hyponymy} & ENGLISH~cousin\penalty-10000 the child of your aunt or uncle & CHINESE~\textchinese{堂妹}\penalty-10000 younger female patrilineal cousin \\\hline
%\textbf{hypernymy} & cugina$_\text{ ITA}$ & female cousin & cousin$_\text{ ENG}$ & the child of your aunt or uncle \\\hline
\end{tabular}
\caption{\label{linguistic_mappings}Examples for interlingual mapping types, used as test cases in the paper, in Malayalam, Tamil, Chinese, and English.}
\end{center}
\end{table}

The paper is organized as follows.
Section~2 provides the theoretical background. Section~3 presents the mapping models of five state-of-the-art exemplary MLDBs. Section~4 provides a quantitative evaluation of a set of relevant reference resources, as well as a comparison of their mapping models.
%Section~5 presents the diversity-aware mapping approach of the UKC, addressing the shortcomings of existing systems. 
Finally, Section~5 provides the conclusion. %future research directions t diversity-awareness in multilingual resources. 
Throughout the paper we will use the example of family relationships---well known to be expressed in diverse manners across languages \cite{khishigsuren2022using}---and in particular the notion of \emph{cousin}, in nine languages: English, French, Italian, Chinese, Hindi, Tamil, Malayalam, Hungarian, and Mongolian.\footnote{The English \emph{cousin} does not have a precise equivalent in six out of the eight other languages.}

\section{Cross-Lingual Lexical Mappings}
\label{problem}

Lexical equivalence is understood by linguists as a complex and multidimensional problem, ranging from multiple coexisting forms of meaning equivalence \cite{adamska2010examining} to  \emph{untranslatability} \cite{catford1978linguistic} (see Table~\ref{linguistic_mappings} for examples). %Often, acceptable translations can be devised through a free combination of words, such as ***. %Within the scope of lexical semantics, however, such cases register as \emph{lexical gaps} for lack of a concise representation in the target language \cite{***}.
While the latter phenomenon, i.e.~the absence of certain lexical mappings, cannot be entirely explained through systematic principles \cite{lehrer1970notes}, differences from one language to another are often due to diversity in culture or the reality perceived. Some examples are: the lack of vocabulary for sailing in Mongolian, the language of a landlocked country, the Italian word \emph{malga} meaning a kind of mountain restaurant, the Scottish Gaelic \emph{onfhadh} meaning \emph{the raging sound of the sea}, or the rich East Asian vocabulary on the various forms of \emph{rice} as grain and as food.
%
% REMOVED FROM EACL : the Hungarian \emph{láb} designates the entire lower limb whereas English only has words for its specific parts: \emph{foot}, \emph{leg}, etc.), 
%

At the same time, traditional bilingual dictionaries remain pragmatically-built and practice-oriented tools for the general public, typically lacking a fine-grained and theoretically precise modelling of the cross-lingual mapping of meaning \cite{ten2016bilingual}. The relationships provided by dictionaries usually imply a quasi-equivalence of word meanings or, more rarely, a broader target meaning if the target language does not have a close enough word sense. Some dictionaries also indicate \emph{lexical gaps}, i.e.~where the target language does not lexicalize the meaning of the source word, as free-text definitions. 
Furthermore, bilingual dictionaries have always been designed to be asymmetric, clearly defining the source and the target language, and the reverse counterpart is never constructed by the mere inversion of its entries. This is due to translation, even when applied to individual word senses, being by nature 
asymmetric and intransitive \cite{adamska2010examining}.
In the context of MLDBs, however, the principle of asymmetry is never respected in practice, for reasons of scalability: if a MLDB supports $n$ languages then mappings would need to be defined for $n(n-1)$ language pairs. In order to reduce the number of mappings needed, all MLDBs rely on a \emph{hub} (or \emph{pivot}) meaning representation to which all lexicons are mapped. The possibility of a hub meaning~$c$, however, is based on the simplifying assumption that the mapping of word meanings is an equivalence relation that, by definition, is symmetric and transitive: $$ m_a\leftrightarrow c\leftrightarrow m_b \Rightarrow m_a\leftrightarrow m_b. $$ For this reason, MLDBs tend to rely mostly on equivalence mappings and, instead, express broader--narrower relationships either within their hub or within language-specific lexicons.

The observation above motivates our goal of comparing the cross-lingual semantic expressivity of MLDBs. The first and fundamental evaluation criterion relates to lexical concepts: it is the ability of the MLDB to represent language-specific lexical meaning. When the hub meaning space of an MLDB is limited to that of a particular language (such as English), it means that the entire database is biased towards that language, as certain lexicons cannot be represented with the same level of detail as others. Beyond the space of meanings, we also evaluate interlingual mapping ability, namely the semantic expressivity of interlingual relations. These should be able to represent interlingual meaning equivalence, but also non-equivalent correspondences and untranslatability, as illustrated in Table~\ref{linguistic_mappings}.

Accordingly, we are going to compare MLDBs with respect to the four criteria below:
%
%These assumptions, as we said before, are not backed up by linguistic theory.%Still, MLDBs take this assumption on the pragmatic grounds of economy and usefulness, a stance that we also adopt in the rest of this paper.

%To sum up, the building of MLDBs, just like of bilingual dictionaries, is and has always been a practical exercise that stays on pragmatic grounds of usability rather than being an attempt at a theoretically sound and exhaustive modelling of language. This means that judging the quality of an MLDB should also stay on quantitative and pragmatic grounds, 
%
\begin{enumerate}
\item \textbf{Unbiased lexical meaning space:} whether the MLDB can represent language-specific lexical concepts for any of the languages it covers, or it is fixed and bound to the meanings from one specific language.
\item \textbf{Interlingual equivalence relation:} whether the MLDB can express concept equivalence for any language pair (among the languages supported).
\item \textbf{Interlingual hypernymy relation:} whether the MLDB can express broader--narrower relationships for any language pair (among the languages supported).
\item \textbf{Untranslatability relation:} whether the MLDB represents lexical gaps as a way explicitly to indicate untranslatability for any language pair (among the languages supported), distinguishing it from the mere absence of a mapping that implies lexicon incompleteness \cite{bentivogli2000looking}.
\end{enumerate}
\noindent
Mapping relations beyond equivalence have major uses in cross-lingual applications. For example, a machine translation (MT) system translating the English sentence \emph{``This rice is tasty''} into Swahili (but also Japanese, Hindi, etc.)  can be informed by an MLDB of the fact that Swahili has no equivalent word for \emph{rice} (untranslatability); instead, it has the more specific words \emph{mchele}, meaning \emph{uncooked rice}, and \emph{wali}, \emph{cooked rice} (hyponymy). This knowledge helps the MT system select the best translation depending on the context, \emph{wali}, and avoid the incorrect \emph{mchele} that leads to a translation with the unintended meaning \emph{``this raw rice is tasty''}. An MLDB that does not distinguish untranslatability from lexicon incompleteness---where an equivalence mapping from \emph{rice} to Swahili is is simply missing---will not be able to inform the MT system of the difficulty within the sentence, and a purely corpus-statistics-based approach may lead to erroneous translation, even in state-of-the-art systems such as Google Translate.\footnote{Google translates the sentence above into the Swahili \emph{``Mchele huu ni kitamu.''}. Such semantic mistakes are frequent in all major translator tools as of today.} 
To our knowledge, the three kinds of interlingual relationships covered by our criteria are on a par with interlingual mappings provided in the best traditional bilingual dictionaries. While in principle we could consider other types of associative cross-lingual relations, such as etymology or cognacy, most MLDBs reviewed in this paper do not contain such information and thus they would not be useful for purposes of comparison.

%We provide a qualitative evaluations of the four criteria above in Section~3, based on existing publications and evidence from resource content. In order to complement these results with quantified data on what the presence or the lack of an ability means in terms of expressivity, in Section~4 we provide a quantitative evaluation on the coverage of interlingual mappings theoretically achievable by each MLDB, based on an expert-created gold standard. 

Throughout the paper we will use the running example of family relationships---well known to be expressed in diverse manners across languages---and in particular the notion of being the \emph{cousin} of somebody, in nine languages: English, French, Italian, Chinese, Hindi, Tamil, Malayalam, Hungarian, and Mongolian. The English \emph{cousin} does not have a precise equivalent in six out of the eight other languages. Instead, they lexicalize more specific concepts among the no less than 63 combinations of \emph{the elder-younger son-daughter of my father's-mother's elder-younger brother-sister}. Thus, in French and Italian, distinct words (inflections) exist to represent the female cousin (\emph{cousin/cousine} and \emph{cugino/cugina}). In Chinese, eight words express the elder-younger son-daughter of your mother's-father's sibling (\textchinese{表姐; 表妹; 表哥; 表弟; 堂姐; 堂妹; 堂兄; 堂弟}). Hindi also uses eight distinct words, yet they are not equivalent to the Chinese ones: they express the son-daughter of your mother's-father's brother-sister (\texthindi{फुफेरा भाई; चचेरा भाई; ममेरा भाई; मौसेरा भाई; चचेरी बहन; फुफेरा बहिन; मौसेरा बहिन; ममेरा बहिन}). Malayalam and Tamil, finally, each have no less than 16~distinct words to express the elder-younger son-daughter of your mother's-father's brother-sister. % (\textmalayalam{പെങ്ങള്; അനുജത്തി; ചേച്ചി; അനിയത്തി; മദനി; നാത്തൂൻ; മുറപ്പെണ്ണ്; ഏട്ടത്തി; അണ്ണൻ; അനുജൻ; ആങ്ങള; അനിയൻ; മച്ചുനൻ; അളിയൻ; മുറച്ചെറുക്കൻ; ചേട്ടൻ}).
Examples such as these cannot be ignored as corner cases. In many societies (such as in Southern India) it is a requisite of appropriate communication to express family relations precisely, and fuzziness is culturally not acceptable. Translators, whether human or AI-based, therefore need to deal with such cases in a correct and coherent manner. While translating any of the specific Chinese, Hindi, or Malayalam words into the more general \emph{cousin} is formally correct (even though information is lost), in the reverse direction a non-semantically-motivated (random or corpus-frequency-based) selection among candidate meanings is likely to inject unintended meaning.

\section{Qualitative Analysis}
\label{soa}

Several past and ongoing efforts exist for building lexical resources, with different underlying motivations, solutions, and sizes \cite{gurevych2016linked}.
Among these, our paper addresses resources that:
\begin{itemize}
    \item are \emph{multilingual}, as the focus of our study is the interlingual mapping of lexical meaning;
    \item have a \emph{public and well-defined model} of lexical meaning that makes it possible to perform a formal analysis of lexical expressivity;
    \item \emph{target natural languages}, as cross-lingual practices around specialized (domain) terminology and encyclopedic knowledge are different from general language and are out of scope for this work.
\end{itemize}
Thus, we do not consider in our study otherwise remarkable resources such as Wiktionary\footnote{\url{http://www.wiktionary.org}} (as it is lacking a formal representation of lexical meaning, a model for meaning-based interlingual mapping and, more generally, a formal structure), Glosbe\footnote{\url{http://www.glosbe.com}} or PanLex\footnote{\url{http://www.panlex.org}} (as their internal representation of meaning is not fully public), DBpedia\footnote{\url{http://www.dbpedia.org}} or ConceptNet\footnote{\url{https://conceptnet.io}} as they are encyclopedic rather than lexical databases. Nor do we consider terminologies such as Agrovoc\footnote{\url{https://agrovoc.fao.org}} as phenomena of linguistic diversity within specialised vocabularies is not the topic of our research. 

We review and compare \emph{EuroWordNet}, \emph{BalkaNet}, the \emph{Multilingual Central Repository}, two versions of the \emph{Open Multilingual Wordnet}, \emph{IndoWordNet},  \emph{BabelNet}, and the \emph{Universal Knowledge Core}, showing how they take markedly different approaches to modelling cross-lingual mappings. Each review consists of a structural overview and an analysis of mapping ability based on a complex example of interlingual mappings around cousin-like family relationships. Table~\ref{tab:comparison} provides a summary comparison according to the four criteria defined in Section~\ref{problem}. 
%We leave aside resources geared towards encyclopedic or ontological knowledge (e.g.~DBpedia, ConceptNet), those with a not fully formalized or not fully public underlying model (e.g.~Wiktionary, PanLex), word vector models that represent word meanings implicitly (collapsing different meanings into a single word vector), and also the countless domain terminology bases, as the mapping of specialized terms is a very different kind of problem. We should signal whenever they employ similar mappings as one of those reviewed.

All MLDBs studied formally distinguish between \emph{words} and \emph{word meanings}, as the correspondence between the two is often one-to-many (polysemy) or many-to-one (synonymy). For a coherent representation of different MLDBs, in the rest of the paper we adopt the \emph{WordNet} model of word meanings and the corresponding terminology, introduced by \cite{miller1998wordnet,fellbaum2007connecting} and today used in thousands of wordnets and similar resources. In wordnets, lexemes are called \emph{words} (even for multi\-word expressions). A word with a specific meaning is called a \emph{sense}. The senses of synonymous words are linked to a single \emph{synset} (synonym set) that formally represents the synonymous senses as collapsed into a single node. Synsets are interconnected into a graph through hierarchical relations of (intra-lingual) hypernymy and hyponymy (broader and narrower meaning), as in traditional thesauri.

% Based on existing practice within MLDBs, we classify inter-lingual mapping relations of meaning along two dimensions: \emph{cardinality} (one-to-one, one-to-zero, one-to-many) and \emph{relation type} (equivalence, hypernymy, hyponymy).

\begin{table}[t]
    \setlength{\tabcolsep}{3pt}
    \small
    \centering
    \begin{tabular}{l|ccccc}
        \hline
        \textbf{MLDB} & \textbf{Unbiased} & \textbf{Equivalence} & \textbf{Hypernymy} & \textbf{Untranslatability} \\\hline
        EuroWordNet & No & Partial & Partial & Partial\\
        BalkaNet & No & Partial & Partial & Partial\\
        MCR & No & Partial & Partial & No\\
        OMW & No & Partial & Partial & No\\
        OMW2 & Yes & Yes & Partial & Yes\\
        IndoWordNet & Yes & Partial & Partial & No\\
        BabelNet & Yes & Yes & Yes & No\\
        UKC & Yes & Yes & Yes & Yes\\\hline
    \end{tabular}
    \caption{Comparison of the support of interlingual meaning representation and mapping features among MLDBs, as defined in Section~\ref{problem}.}
    \label{tab:comparison}
\end{table}

\subsection{EuroWordNet, BalkaNet, MCR, Open Multilingual Wordnet v1~\&~v2}
\label{sec:omw}

% \begin{figure*}
% \centering
% \includegraphics[width=14cm]{f21.png}
% \caption{Example of English - Chinese mapping meaning in OMW} \label{fig7}
% \end{figure*}

% \begin{figure*}
% \centering
% \includegraphics[width=14cm]{f34.png}
% \caption{Example of English - Chinese mapping meaning in GWG} \label{fig7}
% \end{figure*}

\begin{figure*}[t]
\centering
\includegraphics[width=12cm,height=9cm]{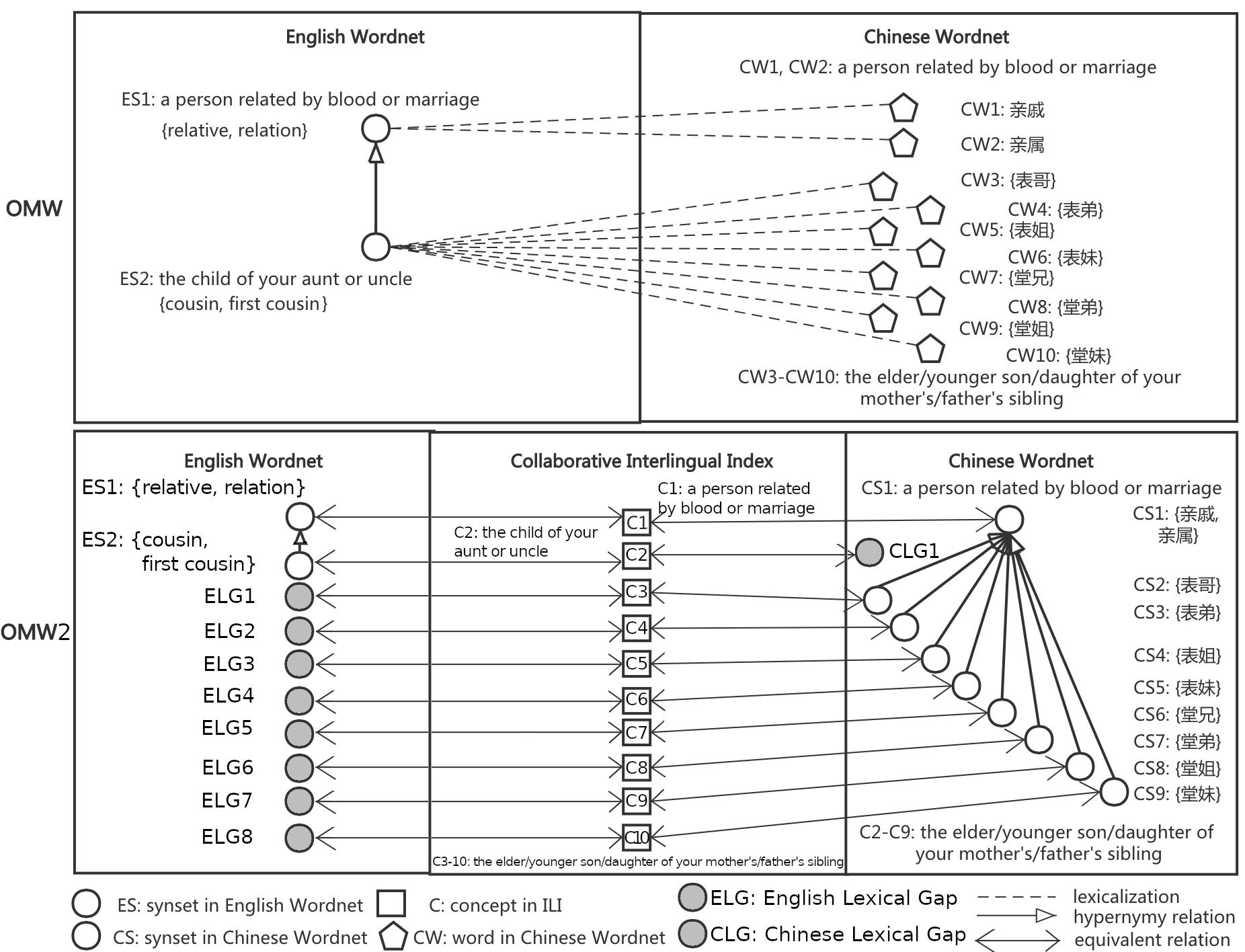}
\vspace{-.1cm}
\caption{Examples of Chinese-to-English mappings using the EuroWordNet/BalkaNet/OMW data models (top) and OMW2 (bottom).} \label{gwg}
\vspace{-.3cm}
\end{figure*}

Due to the many shared features, this section describes together EuroWordNet (EWN) \cite{vossen1997multilingual,vossen1998introduction}, BalkaNet \cite{tufis2004balkanet}, the Multilingual Central Repository (MCR) \cite{gonzalez-agirre2012own}, as well as two versions of the \emph{Open Multilingual Wordnet} (OMW and OMW2) \cite{bond2012survey,bond2013linking,bond2020some}. The EuroWordNet project pioneered the creation of multilingual wordnet resources and their cross-lingual mappings. It directly or indirectly influenced other collaborative efforts, under the umbrella of the \emph{Global WordNet Association} \cite{vossen2016toward,pease2008building},\footnote{\url{http://globalwordnet.org/}} on specific language groups such as BalkaNet for the Balkans and MCR for the languages of Spain. The OMW, in turn, harmonised the representations of these and many other wordnets, e.g.~\cite{black2006introducing,balkova2004russian}, mapped all of them to the English Princeton WordNet~3.0 and, in its \emph{Extended} version, expanded linguistic coverage to hundreds of languages with words automatically extracted from Wiktionary and the \emph{Unicode Common Locale Data Repository}.

All of these efforts use the English Princeton Wordnet (PWN) as their inter-lingual hub. EuroWordNet and BalkaNet link the synsets of separate language-specific wordnets to English PWN synsets through equivalence relations. MCR and OMW, on the other hand, link English synsets directly to words in other languages through \emph{lexicalization} relations that, in practice, still imply meaning equivalence.
In both cases, the use of PWN as a hub results in a bias towards the English language and culture: our criterion~1 on an unbiased meaning space is not fulfilled. Accordingly, MCR and OMW do not contain any word that has no equivalent English meaning in PWN. Some wordnets from EWN (e.g.~Dutch) and BalkaNet (e.g.~Romanian, Czech) contain language-specific synsets and lexical gaps, but the synsets are not mapped to other languages and the gaps are only mapped to English (hence the ``partial'' support for untranslatability in Table~\ref{tab:comparison}).

Figure~\ref{gwg} shows an example of Chinese-to-English mapping in OMW (the EWN/MCR/BalkaNet models behave the same way). The Chinese word CW1 is correctly mapped to the English meaning ES1 \{\emph{relative, relation}\}. The eight Chinese words representing cousins are, however, all mapped to the single PWN synset meaning \emph{cousin}. This results in a representation that is both incomplete and incorrect: the meanings of the more specific Chinese words are lost, while the mappings give the impression that these words are all synonyms and equivalent in meaning to the English \emph{cousin}. The fact that these resources cannot express that the Chinese terms are more specific than \emph{cousin} means that our criterion~3 on hypernymy is only partially fulfilled. Likewise, neither equivalence not untranslatability can be expressed for meanings not present in English (such as the ones in Table~\ref{linguistic_mappings}).

Recently, efforts towards a second version of OMW were announced \cite{bond2020some}. Despite the fact that, as of early 2023, no data has been released for OMW2, we review the abilities of this database based on information available from the publications cited.
OMW2 replaces the lexicalisation mappings of OMW (that relate English PWN synsets with lexicalisations from other languages) by synset-to-synset mapping relations towards a \emph{Collaborative Interlingual Index} (CILI). The CILI is a set (i.e.~an unstructured collection) of unique IDs that represent word meanings relevant to one or more languages. IDs within the CILI are linked to synsets within wordnets with one-to-one equivalence relations (implemented as \texttt{owl:sameAs} in the Semantic Web representation of the OMW2). The collaboratively-built and managed CILI is meant to expand beyond PWN to cover synsets that have no English equivalents, and thus eliminate the English-centeredness of OMW. OMW2 also introduces lexical gaps in order to distinguish between resource incompleteness and untranslatability.

Figure~\ref{gwg} shows the same \emph{cousin} example as it can be modeled by OMW2. It allows the creation of new IDs within the CILI for the eight specific kinds of Chinese cousins, which can then be linked to other languages, or represented as lexical gaps. The eight Chinese meanings can thus be included in the CILI and their absence from the English vocabulary can be explicitly marked. Criteria~1, 2, and~4 (on the unbiased meaning space, equivalence, and untranslatability) are thus fulfilled.
%
%However, there is another problem when mapping the meaning ES2 in English to the corresponding Chinese meanings: According to the semantic relations through English and Chinese wordnet, the GWG system can only deduce that C2 and C3-C10 are at the same level, both are the hyponym concepts of C1, however,C3-C10 is the hyponym concept of C2. As a result,it is hard for GWG to determine the hypernymy relation between ES2 and CS2-CS9 and it can not realize this one-to-many mapping.
%
Note, however, that the graph in Figure~\ref{gwg}, composed of hypernymy edges within the wordnets as well as of equivalence relations towards the CILI, does not provide any relationship between the English meaning of \emph{cousin} (ES2) and the more specific Chinese words (CS2--CS9). The fact that \emph{cousin} is more general than CS2--CS9 is an example of \emph{interlingual knowledge} that is not directly derivable from the union of monolingual lexicons and the CILI.  Even if one wanted to represent this knowledge, it would not be possible within the OMW2 model using the CILI and equivalence mappings alone. As the CILI layer leaves hierarchical structuring of word meanings to individual wordnets, it cannot express cross-lingual hierarchical relationships. Criterion~3 on interlingual hypernymy is therefore only partially fulfilled. 

\begin{figure*}[t]
\centering
\includegraphics[width=12cm,height=5cm]{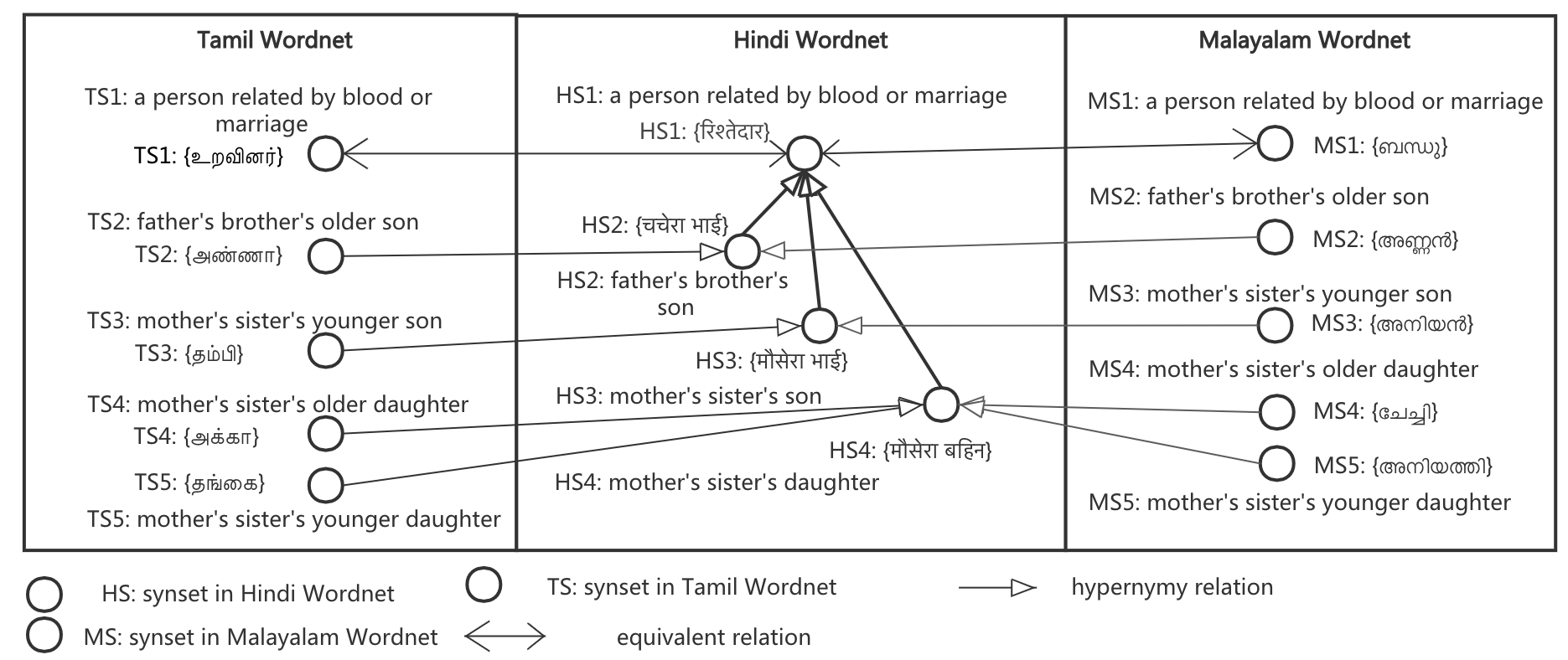}
\vspace{-.2cm}
\caption{Example of Tamil--Hindi--Malayalam mappings using the IndoWordNet model.} \label{iwn}
\vspace{-.3cm}
\end{figure*}

\subsection{IndoWordNet}

\emph{IndoWordNet}\footnote{%https://www.cfilt.iitb.ac.in/indowordnet/,\\ 
https://tdil-dc.in/indowordnet/} (IWN) includes 18~languages from the Indo-Aryan, Dravidian, and Sino-Tibetan families \cite{dash2017wordnet,bhattacharyya2010indowordnet,singh2016mapping,kanojia2018indian,saraswati2010hindi}.
Similarly to other wordnets, IWN uses synsets to represent word meanings along with their associated glosses. One of the particularities of IWN is its use of the \emph{Hindi WordNet} (HWN) \cite{narayan2002experience,chakrabarti2004creation}, as opposed to English, as the central hub that interconnects the 18~languages. %The HWN was created manually using existing dictionaries, and other languages were mapped to it through \emph{expansion}, i.e.~synset-by-synset translation from HWN \cite{bhattacharyya2010indowordnet}. Mappings from IWN towards PWN are also under development \cite{kanojia2018indian,nair2019aligning}. 
Within IWN, only the HWN contains a synset hierarchy: the other 17~languages are represented as flat lists of synsets. The use of HWN (as opposed to PWN) as the hub makes sense for reasons of cultural and linguistic proximity to other languages of India. Accordingly, the HWN contains many synsets culturally and linguistically relevant to the Indian subcontinent.

While the limitation of word meanings to what is lexicalized in Hindi restricts the expressivity of IWN, the database does allow the creation of synsets specific to each of its 17~languages covered. Thus, IWN fulfils our criterion~1 on having an unbiased meaning space. However, such language-specific meanings are not part of the hub which is limited to Hindi. Interlingual equivalence mappings therefore are limited to what is expressed by the Hindi lexicon.

This limitation is counterbalanced by the ability of IWN---unique among the resources reviewed---to use both equivalence and hypermymy for interlingual mapping. Figure~\ref{iwn} shows our \emph{cousin} mappings between Hindi and Malayalam, a Dravidian language from Southern India. In Malayalam, MS1 can be mapped to HS1 using equivalent mapping, but MS2--MS17 are more specific meanings than HS2--HS9 which do not exist in HWN.
The solution of IWN is to link them to a more general synset with hypernymy relations: it maps HS2 (father's sister's son) in Hindi to two more specific Malayalam meanings, MS2 and MS3 (father's sister's elder/younger son) through two hypernymy relations. IWN is thus capable of correctly mapping non-equivalent synsets across languages. On the other hand, due to Hindi being the hub, IWN is not able to map equivalent meanings across Indian languages if the meaning is not part of Hindi. For example, Tamil and Malayalam have lexicalizations for \emph{mother's sister's elder daughter} (TS4 and MS4, resp.), but the IWN can only indicate that they are both hyponyms of HS4, resulting in information loss. IWN thus only partially fulfils criteria~2 and~3 on unbiased equivalence and hypernymy mappings. Finally, the lack of modelling lexical gaps means that IWN fails our criterion~4 on untranslatability.

\begin{figure*}[t]
\centering
\includegraphics[width=12cm,height=6.5cm]{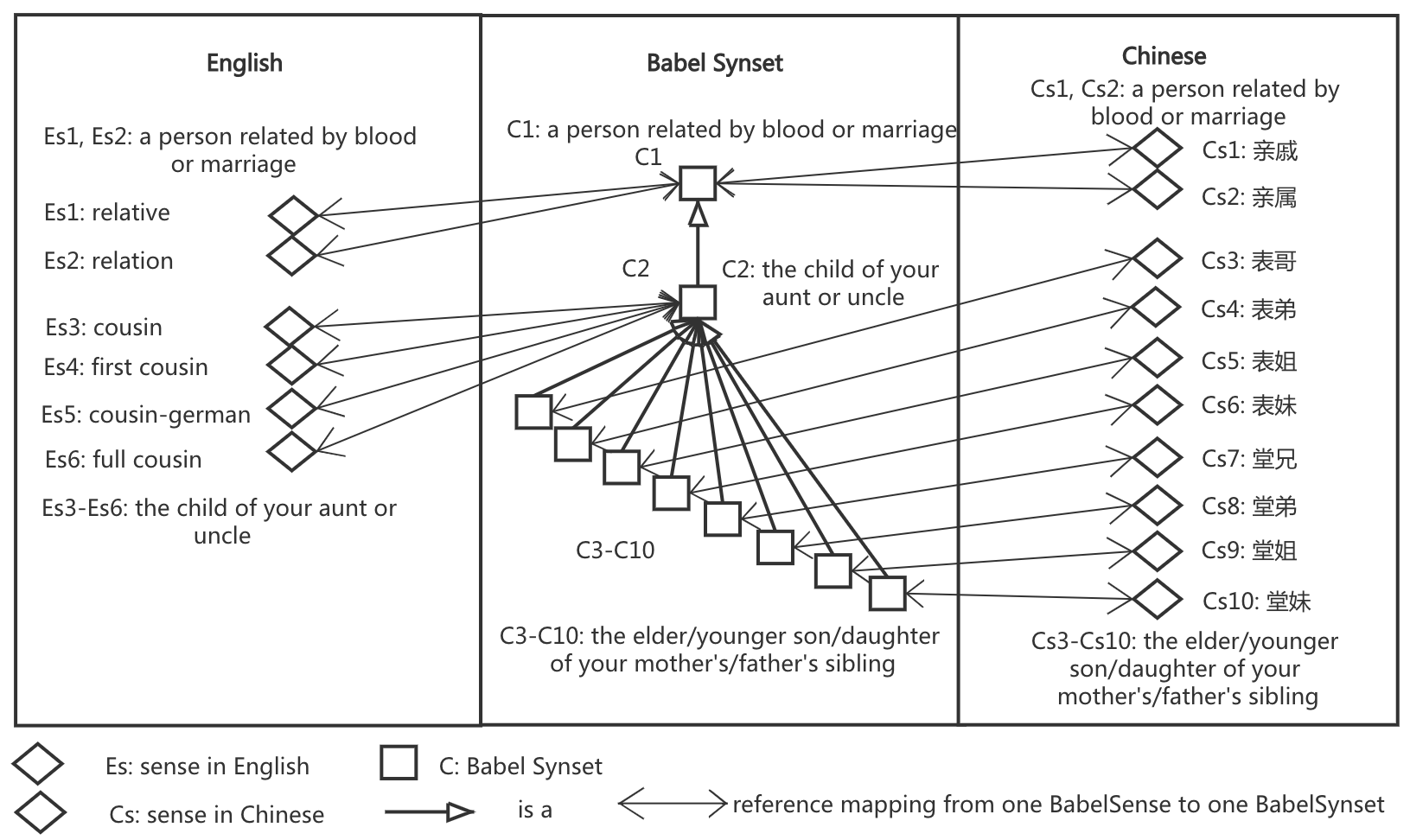}
\vspace{-3mm}
\caption{Chinese-to-English mappings using the BabelNet model.} \label{bn}
\vspace{-3mm}
\end{figure*}

\subsection{BabelNet}

BabelNet\footnote{https://babelnet.org/} stands between a semantic network and a lexical database, covering terms of both lexicographic and encyclopaedic origin \cite{Roberto2012babelnet,EHRMANN14.810}. Version 5.2 of BabetNet contains 520~languages, and 22~million entries. Its contents were imported from online encyclopaedias and lexical resources such as wordnets, Wiktionary, Wikipedia, OmegaWiki, and Wikidata, which explains its larger size and wide coverage of named entities.

% BabelNet was created by integrating multilingual Web encyclopedia, such as Wikipedia, OmegaWiki, Wikidata, Wiktionary and dozens of wordnets.

BabelNet builds a unified, supra-lingual lexical meaning space, represented as a hierarchy of \emph{BabelSynsets}. These, in turn, are lexicalized in each language by language-specific \emph{BabelSenses}. As the synset hierarchy is defined outside of the language-specific lexicons, it becomes theoretically possible to build a meaning space unbiased towards any particular language.
%Compared with OMW and IWN, the semantic graph can maintain the superset of concepts in diverse language cultures, which makes the resource more neutral to different languages. And compared with GWG,it makes the whole semantic space independent of the lexical resources of each specific language, which can further eliminate mismatching between semantic spaces of language resource.
%
Figure~\ref{bn} shows how our running example of English--Chinese mappings could in theory be represented in BabelNet. The supra-lingual central layer is capable of representing shared meanings (e.g.~C1) as well as language-specific meanings (C2--C10), within a single hierarchy. Individual BabelSynsets are then mapped to one or more synonymous lexicalisations (BabelSenses) in each language. The model of BabelNet thus allows word meanings to be hierarchically related across languages (such as the English \emph{cousin} and the eight more specific Chinese meanings), which is not possible for the DBs described in Section~\ref{sec:omw}. It also avoids the limitation of IWN of not being able to map meanings that are not in the hub language. BabelNet thus fulfils criteria~1 to~3, but not criterion~4 as it does not offer any information on untranslatability.

In practice, however, BabelNet does not exploit its structural potential to address language diversity explicitly. %, focusing instead on high encyclopedic coverage. While its structure itself is capable of modelling a richer set of cross-lingual mappings than other state-of-the-art MLDBs, its actual contents are still biased.
This becomes clear by observing how BabelNet 
 actually represents the eight Chinese meanings CS3--CS10: in contrast to the correct mappings shown in Figure~\ref{bn}, it maps most of them to the PWN meaning of \emph{cousin} and leaves the remaining ones unmapped.

\subsection{The Universal Knowledge Core}
\label{ukc}

\begin{figure*}[t]
\centering
\includegraphics[width=12cm,height=6.5cm]{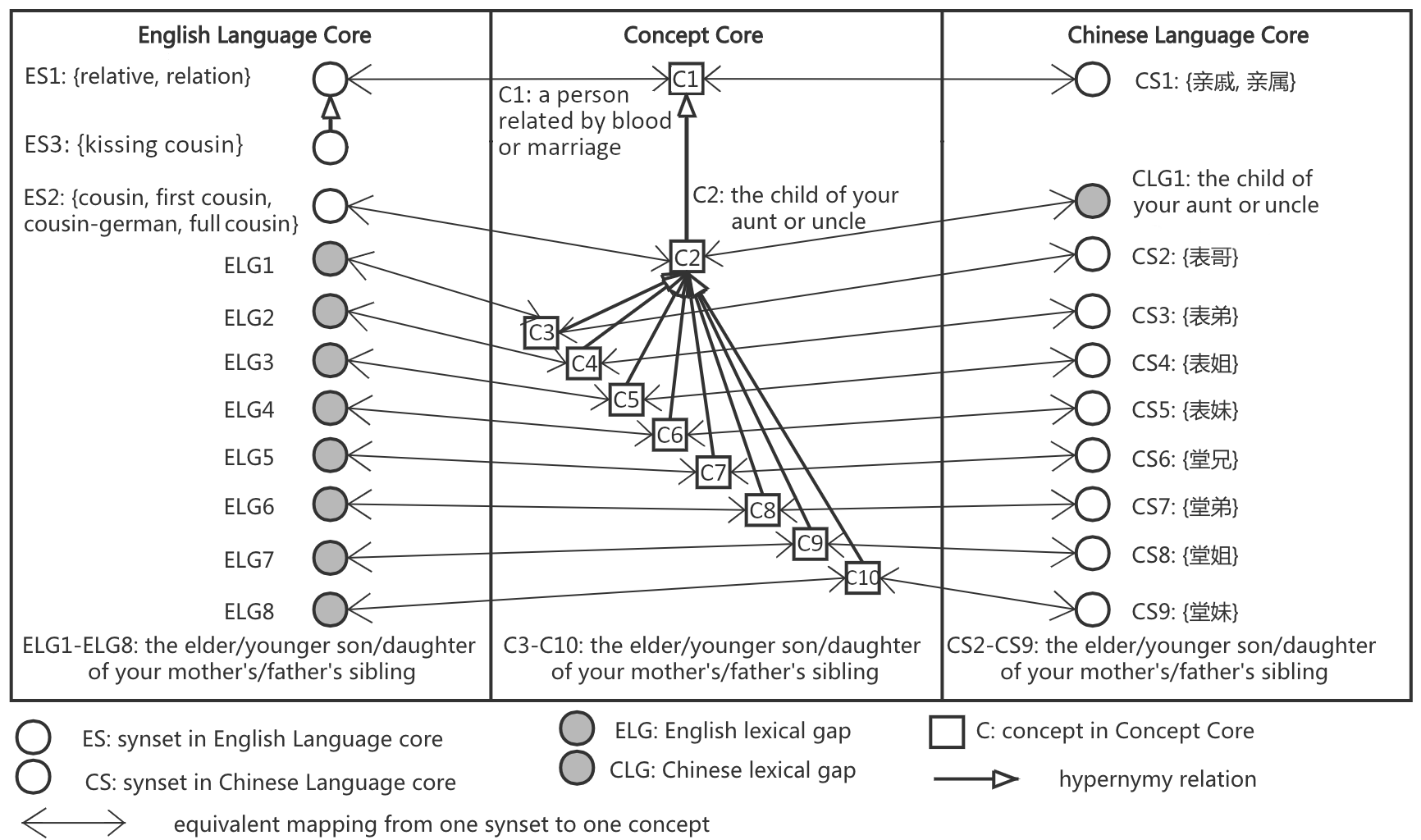}
\vspace{-0.2cm}
\caption{\label{fig7}Example of English--Chinese mappings as supported by the UKC}
\vspace{-0.4cm}
\end{figure*}

The Universal Knowledge Core (UKC) \cite{giunchiglia2017understanding,giunchiglia2018one} is a large-scale MLDB that contains about 2~million words in over 2,000 languages \cite{bella2022language}.\footnote{http://ukc.datascientia.eu/} It integrates a variety of resources such as individual wordnets such as \cite{ganbold2018using, bella2020major}, Wiktionary, as well as original multilingual content on phenomena related to linguistic diversity \cite{giunchiglia2017understanding}, such as cognacy \cite{batsuren2022large}, metonymy \cite{khishigsuren2022metonymy}, lexical gaps \cite{khishigsuren2022using}, morphology \cite{batsuren2021morphynet},
lexical similarity \cite{bella2021database}. %Its main focus being lexical diversity, it was designed to solve the structural coverage and bias issues in cross-lingual mappings observed in other MLDBs. Thus, the structure of the UKC allows the full and entirely bias-free coverage of cross-lingual mappings in our evaluation set.
The UKC has a two-layered architecture, with a \emph{language layer} that contains a separate wordnet-like graph (with words, senses, and synsets) for each language, as well as a supra-lingual layer of \emph{interlingual concepts} \cite{giunchiglia2018one} (Figure~\ref{fig7}). Each such concept represents a word meaning from at least two of the constituting languages, so that the concept layer consists of the \emph{union} of all word meanings that are mapped to at least one other language. Thus, in our running example, each of the eight Chinese meanings of \emph{cousin}, the eight Hindi meanings, and the 16~Malayalam meanings becomes a separate interlingual concept. The UKC thus has an unbiased meaning space (criterion~1).

Yet, the UKC does not assume that lexical meaning within all languages can be perfectly described with a single unified concept graph. A major distinguishing feature with respect to all previously presented MLDBs is the ability to represent word meanings and their hierarchy \emph{both} on the interlingual and on the language-specific levels, the former using concepts and the latter synsets. Thus, we allow smaller unaligned hierarchies to coexist with the merged core of interlingual meanings. This architectural choice reacts to the impossibility of ever reaching a perfect merge of \emph{all} lexicons for \emph{all} languages of the world, both due to the effort implied and allowing for irreducible cases of diversity. For example, in Figure~\ref{ukc}, the newly introduced culture-specific English \emph{kissing cousin}, meaning \emph{a relative with whom someone is in kissing terms}, may need to be aligned with concepts from other languages before it can be integrated into the concept layer, and is thus temporarily kept as a synset-level meaning within the English language layer, all the while being linked to concepts in the overall UKC graph through hypernymy.

\emph{Interlingual equivalence} is represented in the UKC by mapping language-specific synsets to the same concept. For example, the UKC maps the English synset \{\emph{relative, relation}\}, the Italian \{\emph{parente, familiare}\}, and the Chinese \{\textchinese{亲戚,亲属}\} to the same interlingual concept. Thus, the interlingual concept layer acts as the hub and the UKC, just like OMW2, is capable of representing equivalence mappings (criterion~2).

\emph{Interlingual hypernymy and hyponymy} are represented within the concept layer. In this respect, the UKC is different from OMW2 which keeps meaning hierarchies within the original resources. Representing all word meanings as well as their relationships in a single graph means that, as in the case of BabelNet, any pair of word meanings can be put in a broader--narrower relation (criterion~3). 

\emph{Untranslatability}, finally, has explicit support in the UKC through the \emph{lexical gap synset} that, contrary to regular synsets, does not have senses or words attached to it, but does have a gloss. When a concept is not lexicalized in a language, it is mapped to a lexical gap synset instead of leaving it unmapped (as shown in Figure~\ref{fig7}). This feature allows for distinguishing resource incompleteness from untranslatability (criterion~4).
%
%The combination of the three relationship types further increases representational power. If a concept~$c$ is lexicalized in language~$l_A$ as $w_A$ but is a lexical gap in $l_B$, while its parent concept $c_p$ is a gap in $l_A$ but is lexicalized in $l_B$ as $w_B$, then $w_B$ can be considered as a valid (broader) translation candidate for $w_A$.

The ability of the UKC to represent interlingual equivalence, hypernymy, and untranslatability can be exploited in computational applications such as machine translation or cross-lingual transfer learning, in order to improve their precision in linguistically diverse domains. For example, when translating the Chinese \textchinese{堂妹} (younger female patrilineal cousin) to English, a machine translation system can be informed by the UKC that the Chinese word has no English equivalent (it is a gap in English), but that a broader English word \emph{cousin} exists, which is the most suitable single-word translation available. This operation is not symmetric: \textchinese{堂妹} should not be automatically considered as a correct translation for \emph{cousin}, as it implies additional information that may be wrong depending on the context.

\section{A quantitative Evaluation}

\label{sec:comparison}

\begin{table}[t]
\begin{center}
\small
\setlength{\tabcolsep}{2pt}
\raisebox{0mm}{
\begin{tabular}{lrrrrrr}
\hline
\textbf{Criterion} & \textbf{Gold std.} & \textbf{OMW-like} & \textbf{OMW2} & \textbf{IWN} & \textbf{BbN} & \textbf{UKC} \\\hline\hline
1. Concept coverage & 32 & 56.3\% & 100\% & 100\% & 100\% & 100\% \\\hline
2. Equivalence rel. & 431 & 88.9\% & 100\% & 77.3\% & 100\% & 100\% \\
3. Hyper+hypo rel. & 1,139 & 61.6\% & 80.9\% & 90.1\% & 100\% & 100\% \\
4. Untranslatability rel. & 389 & 0\%    & 100\%    & 0\%    & 0\% & 100\% \\
\textbf{Total relations} & \textbf{1,959} & \textbf{55.4\%} & \textbf{88.9\%} & \textbf{69.4\%} & \textbf{80.1\%} & \textbf{100\%} \\\hline
\end{tabular}
}
% FOR JOURNAL \hspace{5mm}
% FOR JOURNAL \includegraphics[width=8cm]{bias_without_ukc.png}
\caption{\label{results}Interlingual concept and mapping coverage for each MLDB evaluated.
% FOR JOURNAL Left: coverage of mapping types, right: structural bias.
}
\end{center}
\end{table}

We evaluate and compare the MLDBs presented in section~\ref{soa} in terms of our four criteria on interlingual mapping ability: how the structure of each resource determines its \emph{coverage} of language-specific concepts, interlingual equivalence, hyper/hyponymy, and untranslatability mappings. %FOR JOURNAL, as well as its inherent \emph{structural bias} towards the coverage of certain languages.
%
%
%FOR JOURNAL We define the $\text{Bias}(l,D)$ of a database~$D$ towards a language~$l$ as the mapping coverage of~$l$ over all possible relationships~$r$, divided by the arithmetic mean of all~$n$ coverages, centered around zero:
%FOR JOURNAL $$ \text{Bias}(l,D)=\frac{n\times\text{MCvg}(l,D,\forall r)}{\sum_{i=1}^{n}{\text{MCvg}(l_i,D,\forall r)}} - 1$$
%Note that both MCvg and Bias can be interpreted content-wise (based on actual MLDB contents) and structurally (based on the subset of mappings a MLDB is able to implement in theory). In this paper we use the latter structural definition.

\para{Evaluation Data.}
As the focus of this paper are the structural abilities of MLDBs rather than the completeness of their actual content---which varies to a great degree according to the languages covered---we evaluate mapping expressivity on an ad-hoc gold standard set of interlingual mappings.
The dataset consists of $\lvert C\rvert=288$ lexical concepts that include 160~lexicalizations and 128~lexical gaps from nine languages and five phyla (English, French, Italian, Chinese, Hindi, Tamil, Malayalam, Hungarian, and Mongolian), all provided by native speakers. The words were deliberately selected from five culturally diverse semantic groups, belonging to four distinct domains: words expressing various \emph{kinship} relations (siblings, cousins, elder/younger, male/female, etc.), kinds of \emph{watercourses} (according to size), \emph{horses} (male/female, young/adult), and \emph{rice} (raw/cooked, white/brown, cleaned or in the husk). The gold standard set contained the exhaustive mappings within each semantic group, in terms of equivalences, $R_\equiv(C)=431$, hyper/hyponymy, $R_\sqsubset(C)=1139$, and untranslatability, $R_\text{GAP}(C)=389$, totalling in~1,959 gold-standard interlingual mapping relations. The appendix provides the complete list of words and gaps, as well as details on corpus development.

\para{Evaluation method.}
The evaluation consisted of manually analyzing the representational ability of MLDBs against each mapping. We included OMW2, IWN, BabelNet, the UKC, and the OMW, the last one equivalent in its mapping abilities to EWN, MCR, and BalkaNet and thus representative of them as well. This involved the analysis of $1,959\times 5=9,795$ mapping instances.\footnote{As we were interested in the structural properties of IWN, we made abstraction of its limitation to Indian languages.} The appendix gives more detail on how the evaluation of MLDBs was performed against the gold standard corpus.

In order to compute coverage results in Table~\ref{results}, we defined the \emph{interlingual concept coverage} $\text{CCvg}(C,\mathcal{D})$ of an MLDB~$\mathcal{D}$ with respect to a set of lexical concepts~$C$ in the following very simple way:
$$\text{CCvg}(C,\mathcal{D}) = \frac{\lvert C^\mathcal{D}\rvert}{\lvert C\rvert} $$
where $C^\mathcal{D}\subseteq C$ are the concepts from $C$ that $\mathcal{D}$ is able to express. In a similar manner, we defined the \emph{interlingual mapping coverage} $\text{MCvg}(r, C, \mathcal{D})$ of an MLDB~$\mathcal{D}$ with respect to the same set of lexical concepts $C$ and the mapping relation type~$r$ as follows:
$$\text{MCvg}(r,C,\mathcal{D}) = \frac{\lvert R_r^\mathcal{D}(C)\rvert}{\lvert R_r(C)\rvert} \text{ where } R_r^\mathcal{D}(C)\subseteq R_r(C)\subseteq C\times C $$
where $r\in\{\equiv,\sqsubset,\sqsupset,\textrm{GAP}\}$, i.e.~one of the mapping relationships evaluated throughout our paper, $R_r(C)$ is the set of all correct interlingual relations of type~$r$ over the set of concepts~$C$, and $R_r^\mathcal{D}(C)$ is a subset of these relations that $\mathcal{D}$ is able to express.

Quantitative results can be found in Table~\ref{results}. In the following we provide both a discussion of the results.

%\begin{table*}
%\small
%\begin{tabular}{l|l|l|l|l|l|l|l}
%\textbf{Meaning} & \textbf{ENG} & \textbf{FRA} & \textbf{ITA} & \textbf{CHI} & \textbf{HIN} & \textbf{TAM} & \textbf{MAL} \\\hline
%person related by blood or marriage & relative & parent & parente & 亲戚 & \texthindi{रिश्तेदार} & \texttamil{உறவினர்} & \textmalayalam{ബന്ധു} \\
%the child of your aunt or uncle & cousin & cousin & cugino & --- & --- & --- & --- \\
%female cousin & --- & cousine & cugina & --- & --- & --- & --- \\
%elder daughter of your mother's sibling & --- & --- & --- & 表姐 & --- & --- & --- \\
%daughter of your mother's sister & --- & --- & --- & --- & \texthindi{मौसेरा बहिन} & --- & --- \\
%elder daughter of your mother's sister & --- & --- & --- & --- & --- & \texttamil{அக்கா} & \textmalayalam{ചേച്ചി} \\
%\end{tabular}
%\caption{\label{testset}The 16 word senses and 26 lexical gaps used in our evaluations.}
%\vspace{-4mm}
%\end{table*}

\para{Discussion.} All MLDBs evaluated, except for OMW (and the similar EWN, BalkaNet, and MCR), provide a mechanism for adding language-specific concepts to the database. OMW, instead, is limited to the synsets present in the English WordNet, which covers only 18~concepts out of 32~in our gold standard, corresponding to the concept coverage of 56.25\% shown in Table~\ref{results}. 

All MLDBs generally support equivalence mappings and were able to express most of such mappings in our test set. OMW-like databases and IWN, however, are unable to express equivalences that involve meanings that are missing from their hub language (English and Hindi, resp.),
such as \emph{fleuve}$_\textrm{FRENCH}$\,$\equiv$\,\emph{folyam}$_\textrm{HUNGARIAN}$ (meaning a particularly large river) or \emph{mchele}$_\textrm{SWAHILI}$\,$\equiv$\,\textchinese{生米}$_\textrm{CHINESE}$ (meaning uncooked rice). This is a form of structural bias. OMW2, BabelNet, and the UKC, on the other hand, are able to represent all equivalences through their extensible hubs that create a node (a CILI entry, a BabelSynset, and a concept, respectively) for each word meaning lexicalized in at least one language.

In terms of interlingual hypernymy mappings, larger differences are observed among the MLDBs. While BabelNet and the UKC are able to express 100\% of our test set mappings, the remaining resources are weaker. In the case of OMW, EWN, BalkaNet, and MCR, only the PWN-based hub contains a hierarchy, which means that these resources can only express such relations if they are also present in the PWN. Thus, these MLDBs miss 38.4\% of hypernymy and hyponymy from our test set. OMW2 takes the opposite approach and relies on the individual wordnet hierarchies and the cross-lingual equivalence mappings (as shown in Figure~\ref{gwg}) to infer them. This is not sufficient to compute certain mappings, such as the relation between the English \emph{cousin} and the more specific Malayalam words, as the meaning of cousin is a lexical gap in Malayalam. IWN, in turn, is more powerful due to its use of cross-lingual hypernymy mapping relations, and is therefore able to express the English--Malayalam relation (as well as many others) via hypernymy through a Hindi hub meaning. Yet, it would not be able to express hypernymy between \emph{cousin} and the Chinese \textchinese{表姐} as no relation exists between the Chinese and any of the Hindi meanings. BabelNet and the UKC were able to express all mappings as they foresee the creation of a hub concept for each meaning and, contrary to OMW2, define the hierarchy within their hubs.

Finally, for untranslatability mappings, only OMW2 and the UKC provide explicit support for lexical gaps; this is visible from the table within Figure~\ref{results}. All other resources confound gaps with incompleteness, not differentiating a gap from a missing mapping. 

\para{Study limitations.} As stated earlier, our goal  was to quantitatively evaluate the impact of the theoretical mapping abilities of MLDBs on their coverage of a gold-standard interlingual mapping space. The abilities of each MLDB were formalised in our evaluation based on an analysis of their contents (when available) as well as on their descriptions in publications. While we do provide general qualitative information on the actual contents of each MLDB, these contents were not used in our evaluations.

Our evaluation covered concepts taken from four domains well known for their cross-lingual diversity: kinship, animals, geography, and food. We do not expect the inclusion of new diversity-rich domains, such as colors or body parts, to affect our analysis and qualitative findings. That said, a less varied choice of domains and languages (e.g.~the inclusion of more European languages or of lexically more uniform domains such as mathematics) would certainly lead to more homogeneous results in mapping abilities. Our evaluation languages and domains was admittedly and deliberately selected in order to amplify phenomena of lexical diversity as much as possible.

%vspace{1mm}
%\noindent\textbf{Structural bias.}
%Among the MLDBs considered, only the OMW and the IWN have a `hardcoded' preference towards a specific language, namely English and Hindi. This is well reflected in our results (Figure~\ref{results}): the OMW shows a strong bias towards English and the IWN a more moderate bias towards Hindi and Indic languages in general. (Positive values indicate bias \emph{for} a language and negative ones \emph{against} a language). This coincides with negative bias against the five Asian languages for OMW, and against all non-Indian languages for IWN. OMW2 and BabelNet, on the other hand, present only a minimal level of bias: while they are structurally language-agnostic, their inability to express certain mappings results in some cross-cultural data not being represented. Overall, our definition of bias appears to be a meaningful quantitative measure of the language-specificity of a multilingual resource. 

\section{Conclusion}

In this paper we dealt with the problem of how language diversity is represented in state-of-the-art multilingual lexical databases, an important issue in a globalized world where multilingual interactions are the norm and where, at the same time, the vast majority of languages does not benefit from adequate digital support. Current MLDBs should, at the minimum, leave open the possibility for these languages to integrate with the others, all the while avoiding any loss in their capacity of expressing lexical meaning specific to them.
Our analysis, consisting of a theoretical qualitative and an example-based quantitative part, has shown largely differing cross-lingual mapping abilities among the MLDBs examined.%, both in mapping coverage and structural bias. 
We were able to explain these findings by the various ways of MLDBs to define language-specific meaning and their differing support of interlingual mapping.

%As the UKC is intended to be a MLDB that captures lexical diversity, we designed its lexical model to allow a high coverage of language-specific meanings without structural bias towards any language in particular. Support for lexical gaps and merging language-specific meanings into a central concept graph are key techniques for reaching these goals. %At the same time, allowing yet unmerged language-specific graphs to coexist with the central concept graph provides a compromise between the feasibility of a single unified concept graph and cross-lingual mapping ability.

%Providing the structural requirements for supporting language diversity in an unbiased way is only the first step towards building a large-scale MLDB that provies unbiased multilingual content. We believe that our definitions %are also applicable for the evaluation of content, we believe they 
%will be useful tools for quality assurance as part of resource building.

%\newpage
\balance
\bibliographystyle{plain}
\bibliography{manuscript}

\end{document}